\pdfoutput=1

\documentclass[11pt]{article}

\usepackage{acl}

\usepackage{times}
\usepackage{latexsym}
\usepackage{pdflscape}
\usepackage{longtable}
\usepackage{multirow}
\usepackage{color,soul}

\usepackage[T1]{fontenc}

\usepackage[utf8]{inputenc}

\usepackage{microtype}

\usepackage{inconsolata}

\usepackage{graphicx}

\title{Bridging the Gap: Transfer Learning from English PLMs to Malaysian English}

\author{Mohan Raj Chanthran$^1$, Lay-Ki Soon$^{1*}$, Ong Huey Fang$^1$, and Bhawani Selvaretnam$^2$
       \\
       $^1$School of Information Technology, Monash University Malaysia\\
       \{mohan.chanthran, soon.layki, ong.hueyfang\}@monash.edu\\ 
       $^2$Valiantlytix\\
       {bhawani@valiantlytix.com} \\
       }

\graphicspath{ {./figures/} }

\begin{document}
\maketitle

\begin{abstract}
Malaysian English is a low resource creole language, where it carries the elements of Malay, Chinese, and Tamil languages, in addition to Standard English. Named Entity Recognition (NER) models underperform when capturing entities from Malaysian English text due to its distinctive morphosyntactic adaptations, semantic features and code-switching (mixing English and Malay). Considering these gaps, we introduce MENmBERT and MENBERT, a pre-trained language model with contextual understanding, specifically tailored for Malaysian English. We have fine-tuned MENmBERT and MENBERT using manually annotated entities and relations from the Malaysian English News Article (MEN) Dataset. This fine-tuning process allows the PLM to learn representations that capture the nuances of Malaysian English relevant for NER and RE tasks. MENmBERT achieved a 1.52\% and 26.27\% improvement on NER and RE tasks respectively compared to the bert-base-multilingual-cased model. Although the overall performance of NER does not have a significant improvement, our further analysis shows that there is a significant improvement when evaluated by the 12 entity labels. These findings suggest that pre-training language models on language-specific and geographically-focused corpora can be a promising approach for improving NER performance in low-resource settings. The dataset and code published in this paper provide valuable resources for NLP research work focusing on Malaysian English. 

\def\thefootnote{*}\footnotetext{Corresponding Author.}\def\thefootnote{\arabic{footnote}}
\end{abstract}

\section{Introduction}
\label{sec:introduction}

With the recent proliferation of Large Language Models (LLMs), the usage of Pre-trained Language Models (PLMs) like BERT \citep{devlin2019bert}, RoBERTa \citep{liu2019roberta}, XLM-RoBERTa \citep{conneau2020unsupervised}, ALBERT \citep{lan2020albert} has been overshadowed. However, PLM has shown some significant improvements when further pre-trained in domain specific \citep{chalkidis2020legalbert, Lee_2019, araci2019finbert, huang2020clinicalbert} or language specific corpus \citep{Martin_2020, chan2020germans, antoun2021arabert, vamvas2023swissbert}, and subsequently fine-tuned on NLP tasks. The adaptability of pre-trained language model to diverse languages and dialects has enabled its application to specific linguistic contexts, including Malaysian English, a unique and culturally rich variant of the English language.

Malaysian English is also categorized as a creole language due to its distinct characteristics that include loanwords, compound words, and the derivation of new terms from  Malay, Chinese, and Tamil, in addition to the Standard English \citep{chanthran2024malaysian}. Existing state-of-the-art (SOTA) solutions do not produce satisfactory outcomes for downstream tasks performed in Malaysian English \citep{chanthran2024malaysian}. This is mainly due to the morphosyntactic and semantical adaption nature of Malaysian English. Hence, there is a need to improve this SOTA in order to support effective processing of Malaysian English texts.

This work investigates the effectiveness of English pre-trained language models (PLMs) to the low resource language like  Malaysian English, in downstream task, particularly Named Entity Recognition (NER), and Relation Extraction (RE) performed on Malaysian English. Our findings contribute to the growing body of research on promoting inclusivity in NLP by exploring the applicability of PLMs to non-standard English varieties.

The contributions of this paper are as follows:
\begin{enumerate}
    \item Multilingual Pre-trained Model for Malaysian English: We introduce MENmBERT, a BERT-based model pre-trained on a Malaysian English News (MEN) Corpus. MEN Corpus comprises 14,320 articles, facilitating research on applying PLMs to Malaysian English \citep{chanthran2024malaysian}. This model will be made public to develop resources and facilitate NLP research in Malaysian English. The code for this experiment, and dataset have been published in \url{https://github.com/mohanraj-nlp/MEN-Dataset/tree/pretrained-lm}. 
    \item Fine-Tuned and Evaluated NER and RE on Malaysian English: We evaluated the effectiveness of fine-tuned MENmBERT for NER and RE tasks on a benchmark MEN-Dataset. MEN-Dataset contains 200 news articles with 6,061 annotated entities and 4,095 relation instances \citep{chanthran2024malaysian}. This analysis demonstrates the applicability of transfer learning from English PLMs to Malaysian English NLP tasks. 
\end{enumerate}

This paper is structured as follows. Section 2 explores the utilization of pre-trained language models in both English and non-English scenarios. In Section 3, we dive deeper into the pre-training and fine-tune methodologies of MENmBERT and MENBERT. Section 4 we will discuss on the results of fine-tuned NER and RE. Finally, Section 5 concludes the work and shares potential enhancement as the future work.

\section{Related Work}
\label{sec:related-work}

\subsection{Pre-Trained Language Model for Non-English Context}
\label{ssec:pre-trained-language-model-rw}

Language-specific Pre-trained Language Models (PLMs) are essential to handle complex languages and improve performance on downstream NLP task specific to particular language. Considering this, AraBERT has been proposed to address the morphological and syntactic differences in the Arabic language compared to other languages, as Arabic shares very little with Latin-based languages and has unique characteristics \citep{antoun2021arabert}. Multilingual models to learn representations for multiple languages simultaneously resulted in little data representation and small language-specific vocabulary for Arabic, hindering performance compared to a single-language model. \citep{antoun2021arabert} overcome these challenges, the researchers pre-trained AraBERT specifically for the Arabic language to capture the contextualized representations needed for Arabic NLP tasks. By customizing the model for Arabic and optimizing factors such as data size, vocabulary size, and pre-processing techniques, AraBERT was able to achieve state-of-the-art performance on various Arabic NLP tasks. The pre-training has been completed with 70 million sentences and around 24GB of textual Arabic data. AraBERT performs better in downstream tasks like Sentiment Analysis, Named Entity Recognition (NER), and Question Answering \citep{antoun2021arabert}. \citet{antoun2021arabert} compared the AraBERT fine-tuned model to SOTA and M-BERT (also know Multilingual BERT), the results shows AraBERT performing better than mBERT or SOTA.  

Following the success of AraBERT in Arabic NLP, similar approaches can be applied to other low-resource languages with unique characteristics. One such example is KinyaBERT, a recent model specifically designed to address the challenges of Natural Language Processing (NLP) tasks in Kinyarwanda \citep{nzeyimana-niyongabo-rubungo-2022-kinyabert}. KinyaBERT has been implemented with a two-tier BERT architecture that token-level morphology encoder and sentence/document level encoder. By dividing the model's processing into these two tiers, KinyaBERT aims to effectively capture both the fine-grained morphological details of individual tokens and the broader contextual information present in the input text. The pre-training task has been completed with 16 million and 2.4GB of Kinyarwanda language texts. KinyaBERT is evaluated on NLP downstream tasks such as NER, News Categorization Task (NEWS) and Machine-Translated GLUE Benchmark. From the evaluation, KinyaBERT has outperformed the baseline model like BERT Base Pre-trained on Kinyarwanda Corpus (BERT BPE), BERT Tokenized by Morphological Analyzer (BERT MORPHO) and XLM-R. 

Similarly to AraBERT and KinyaBERT, SwissBERT has been proposed specifically for national languages of Switzerland \citep{vamvas2023swissbert}. SwissBERT trained using a combination of domain adaptation, language adaptation, and multilingual approaches. SwissBERT has been fine-tuned for NER task and it was able to outperform the baseline model which has been further pre-trained. SwissBERT has undergone several key adaptations and additions to make it impact for processing Switzerland-related text, this includes:
\begin{enumerate}
    \item Multilingual Adaptation: SwissBERT is trained on a corpus of more than 21 million Swiss news articles in the national languages of Switzerland, including German, French, Italian, and Romansh Grischun.
    \item Custom Language Adapters: SwissBERT utilizes custom language adapters in each layer of the transformer encoder for the four national languages of Switzerland. 
    \item Switzerland-Specific Subword Vocabulary: To further enhance its performance on Switzerland-related text, SwissBERT is equipped with a Switzerland-specific subword vocabulary. 
\end{enumerate}
SwissBERT's superior performance in Switzerland-related tasks, such as Named Entity Recognition and Stance Detection, highlights its efficacy in handling diverse linguistic content specific to Switzerland.

\citet{devlin2019bert} has further pre-trained multilingual BERT (M-BERT) with 104 languages Wikipedia corpus. \citet{wang-etal-2020-extending} suggests enhancing M-BERT by pre-training with low-resource corpora, as 11 languages are not covered in the current 104 languages. The research found that fine-tuning M-BERT (E-MBERT) on low-resource language corpora enhanced NER task performance. \citet{Wu2020AreAL} found that multilingual BERT (mBERT in \citet{Wu2020AreAL}) may not perform well in low-resource languages. This inconsistency may be attributed to the fact that mBERT is trained with a multitude of languages. Pre-training mBERT for a low resource language model can negatively impact the performance compared to training a monolingual BERT model for that language. Findings in related works have inspired us to establish better evaluation and model selection criteria for the development of a pre-trained language model for Malaysian English.


\section{MENmBERT and MENBERT}
\label{sec:men-bert}

\subsection{Overview}
\label{ssec: menbert-intro}

\citet{chalkidis2020legalbert} discussed about two possible further pre-training strategies: 
\begin{enumerate}
    \item Continued / Further Pre-training (FP): FP creates domain- or language-specific BERT models. FP utilises pre-trained model parameters, saving time and data \citep{kalyan2021ammus}. 
    \item Pre-Training from Scratch (SC): Pre-training from scratch lets you train the model with a lot of data. Pre-training the model from scratch will utilise existing language model architecture and parameters \citep{kalyan2021ammus}. 
\end{enumerate}
The difference between two approach is, FP has been pre-trained with generic corpora like BookCorpus, and English Wikipedia \citep{devlin2019bert} while SC has not been pre-trained with any corpus. With this in mind, we proposed to explore several further pre-training strategies:
\begin{enumerate}
    \item MENBERT-FP: Further pre-train bert-base-cased model with MEN-Corpus 
    \item MENmBERT-FP: Further pre-train bert-base-multilingual-cased model with MEN-Corpus
    \item MENBERT-SC: We pre-train bert-base-cased from scratch with MEN-Corpus 
\end{enumerate}
Malaysian English features loan words, compound blend and derivations of new terms from multiple languages local language, making multilingual BERT an effective model for understanding the contexts of news articles. The need to further pre-train BERT, mBERT and train BERT from scratch stems from our hypothesis that PLM with rich language-based understanding will improve the performance of NER and RE after being fine-tuned. Section \ref{ssec: menbert-fp} presents the pre-training setup and hyperparameters used, while Section \ref{ssec: menbert-ft} explains the model fine-tuning tasks. 

\subsection{Further Pre-Training MENmBERT and MENBERT}
\label{ssec: menbert-fp}
Python library Transformers \citep{wolf-etal-2020-transformers} has been used to pre-train BERT. For MENmBERT and MENBERT we have used bert-base-multilingual-cased and bert-base-cased respectively. Since MENBERT-SC was trained from scratch, we used bert-base-cased architecture. We generated our own vocabulary using BertWordPieceTokenizer for MENBERT-SC, meanwhile for MENmBERT and MENBERT we have used BertTokenizer. We selected the hyperparameter combination with the lowest training loss. Table \ref{tab:fp-hp} lists the important hyperparameters used to train the models. Section \ref{sec:experiment-result} details the pre-training results and some analyses. 

\begin{table}[!ht]
\centering
\resizebox{\columnwidth}{!}{%
\begin{tabular}{|c|c|c|c|}
\hline
Hyperparameters       & MENBERT-FP & MENmBERT-FP & MENBERT-SC \\ \hline
epoch                 & 30         & 30          & 30         \\ \hline
batch\_size           & 32         & 16          & 32         \\ \hline
learning\_rate        & 5e-5       & 5e-5        & 5e-5       \\ \hline
weight\_decay         & 0.001      & 0.001       & 0.001      \\ \hline
max\_sequence\_length & 512        & 512         & 512        \\ \hline
\end{tabular}%
}
\caption{Hyperparameters used to train MENBERT-FP, MENmBERT-FP, MENBERT-SC}
\label{tab:fp-hp}
\end{table}

\subsection{Fine-Tuning MENmBERT and MENBERT}
\label{ssec: menbert-ft}

Fine-Tune is an adaptation method to train pre-trained model for any NLP downstream task \citep{kalyan2021ammus}. Three pre-trained model MENBERT-FP, MENmBERT-FP, and MENBERT-SC were fine-tuned for NER and RE using MEN-Dataset. Additionally, we also fine-tuned pre-trained models bert-base-cased and bert-base-multilingual-cased. Fine-tuning on pre-trained and further pre-trained models helps us to compare the performance and validate our hypothesis (see Section \ref{ssec: menbert-intro}).  

\subsubsection{Named Entity Recognition}
\label{sssec: ner-menbert}
We used the Python library Transformers \citet{wolf-etal-2020-transformers}, specifically the BertForTokenClassification module, for fine-tuning. As suggested by \citet{devlin2019bert}, we went through hyperparameter optimization to find an optimal hyperparameter for fine-tuning. We leveraged on WandB \citep{wandb} for hyperparameter optimization and logging. We used [2e-5, 5e-5] for the learning\_rate, [10, 20, 30] for num\_train\_epochs, [0.01, 0.001, 0.0001] for weight\_decay, and finally [4, 8, 16] for the per\_device\_train\_batch\_size.

We used a grid-based search approach to find an optimal hyperparameter with maximum F1-Score and minimum evaluation loss.  Table \ref{tab:optimal-hyperparam-ner} provides the hyperparameters used to fine-tune for NER. The MEN-Dataset is split into training (75\%), test (10\%) and validation (15\%), with total entities of 5065, 453 and 618 respectively. Models fine-tuned with optimal hyperparameter were evaluated using the validation set, and discussed in the following Section \ref{sssec: ft-result-and-analysis-ner}.
\begin{table}[!ht]
\centering
\resizebox{\columnwidth}{!}{%
\begin{tabular}{|c|l|l|l|l|}
\hline
Hyperparameters   & epoch & batch\_size & learning\_rate & weight\_decay \\ \hline
bert-based-cased  & 20    & 4           & 5e-5           & 0.0001        \\ \hline
MENBERT-FP        & 30    & 4           & 5e-5           & 0.01          \\ \hline
mbert-based-cased & 30    & 4           & 5e-5           & 0.01          \\ \hline
MENmBERT-FP       & 30    & 4           & 5e-5           & 0.01          \\ \hline
MENBERT-SC        & 30    & 4           & 5e-5           & 0.01          \\ \hline
\end{tabular}%
}
\caption{Optimal Hyperparameters used to fine-tune bert-base-cased, bert-base-multilingual-cased, MENBERT-FP, MENmBERT-FP, MENBERT-SC for NER}
\label{tab:optimal-hyperparam-ner}
\end{table}

\begin{figure*}[!ht]
    \centering
    \includegraphics[height=\textheight,width=\textwidth,keepaspectratio]{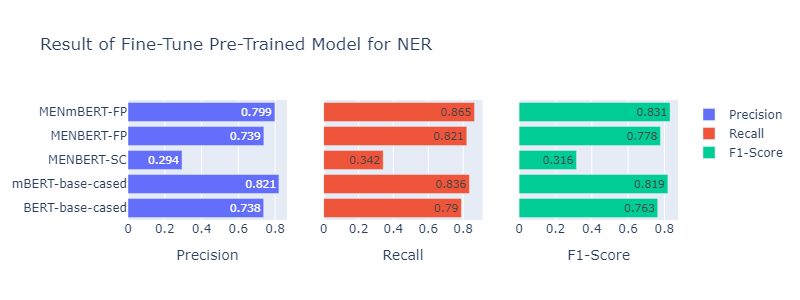}
    \caption{Precision, recall, and F1-score calculated for NER on the MEN-Dataset validation set.}
    \label{fig:Barchart-Evaluation-NER-FineTune}
\end{figure*}

\subsubsection{Relation Extraction}
\label{sssec: re-menbert}

\begin{table}[!ht]
\centering
\resizebox{\columnwidth}{!}{%
\begin{tabular}{|c|l|l|l|l|}
\hline
Hyperparameters   & epoch & batch\_size & learning\_rate & weight\_decay \\ \hline
bert-based-cased  & 30    & 4           & 5e-5           & 0.1        \\ \hline
MENBERT-FP        & 30    & 4           & 5e-5           & 0.1          \\ \hline
mbert-based-cased & 30    & 4           & 5e-5           & 0.1          \\ \hline
MENmBERT-FP       & 30    & 4           & 5e-5           & 0.1          \\ \hline
MENBERT-SC        & 30    & 4           & 5e-5           & 0.1          \\ \hline
\end{tabular}%
}
\caption{Optimal Hyperparameters used to fine-tune bert-base-cased, bert-base-multilingual-cased, MENBERT-FP, MENmBERT-FP, MENBERT-SC for RE}
\label{tab:optimal-hyperparam-re}
\end{table}

To efficiently fine-tune PLM for RE on the MEN-Dataset, we leveraged existing fine-tuning code for document-level relation extraction\footnote{\href{https://github.com/ewrfcas/DocRED_Bert}{DocRED\_Bert Github Link}}. We then carefully modified this code to accommodate the specific characteristics and labeling scheme of the MEN-Dataset. Similarly like NER we went through hyperparameter optimization to find an optimal hyperparameter for fine-tuning. We used [2e-5, 5e-5] for the learning\_rate, [10, 20, 30] for num\_train\_epochs, [0.01, 0.001, 0.0001] for weight\_decay, and finally [4, 8, 16] for the per\_device\_train\_batch\_size. Table \ref{tab:optimal-hyperparam-re} provides the hyperparameters used to fine-tune for RE. 

MEN-Dataset has relation labels adapted from prominent RE dataset like DocRED \citep{yao-etal-2019-docred} and ACE-2005 \citep{Walker2005-ym}. There are 84 relation labels that are adapted from DocRED, and 16 relation labels from ACE-2005. For this study, we concentrated on the relation labels originating from the DocRED dataset. One of the reason for our decision are due to Label Distribution. The DocRED labels constitute the majority within the MEN-Dataset. Focusing on these prevalent labels allows the model to learn robust representations for the most frequently occurring relation types, leading to potentially better performance on tasks involving these relations. Apart from that, we have also excluded a special relation label "NO\_RELATION", as they are used is to indicate entities that might have a relation but not captured by the predefined relation set \citep{chanthran2024malaysian}. Including "NO\_RELATION" could introduce noise or ambiguity during model training. 

MEN-Dataset has a total of 2,237 relation instances adapted from DocRED relation labels, distributed across training (1,693), testing (267), and validation (277) sets. To ensure a comprehensive representation of relation labels during training, we employed a stratified sampling approach on the MEN-Dataset. While the original split allocates 75\%, 10\%, and 15\% for training, validation, and testing, respectively, our stratified sampling guarantees that all relation labels are present in the training data. The result and analysis of fine-tuned PLM for RE have been discussed in Section \ref{sssec: ft-result-and-analysis-re}.

\section{Experiment Result and Analysis}
\label{sec:experiment-result}

\subsection{Further Pre-trained Language Model}
\label{ssec: fp-result-and-analysis}

\begin{table*}[!ht]
\centering
\resizebox{\textwidth}{!}{
\begin{tabular}{|cc|ccccc|}
\hline
\multicolumn{1}{|c|}{\multirow{2}{*}{Masked Sentence}} & \multirow{2}{*}{Masked Token} & \multicolumn{5}{c|}{Pretrained Model} \\ \cline{3-7} 
\multicolumn{1}{|c|}{} &  & \multicolumn{1}{c|}{\begin{tabular}[c]{@{}c@{}}bert-based\\ -cased\end{tabular}} & \multicolumn{1}{c|}{\begin{tabular}[c]{@{}c@{}}mbert-based\\ -cased\end{tabular}} & \multicolumn{1}{c|}{menbert-fp} & \multicolumn{1}{c|}{menbert-sc} & menmbert-fp \\ \hline
\multicolumn{1}{|c|}{\begin{tabular}[c]{@{}c@{}}There are three levels of disaster management, \\ the first involves a locality in a district, \\ secondly when more than two districts of a \\ state is involved and the third involves \\ two or three states. So everyone is aware of this,” \\ he told a press conference at <MASK> Sri \\ Muda today.\end{tabular}} & Taman & \multicolumn{1}{c|}{the} & \multicolumn{1}{c|}{Sri} & \multicolumn{1}{c|}{the} & \multicolumn{1}{c|}{'} & Taman \\ \hline
\multicolumn{1}{|c|}{\begin{tabular}[c]{@{}c@{}}On the Perlindungan Tenang Voucher, \\ he said all eight million recipients \\ of <MASK> Prihatin Rakyat are eligible to \\ receive the voucher worth RM50 announced \\ in Budget 2021 for the benefit of \\ the B40 group.\end{tabular}} & Bantuan & \multicolumn{1}{c|}{the} & \multicolumn{1}{c|}{the} & \multicolumn{1}{c|}{the} & \multicolumn{1}{c|}{the} & Anugerah \\ \hline
\multicolumn{1}{|c|}{\begin{tabular}[c]{@{}c@{}}Ismail Sabri also hoped that ties between \\ Umno and PAS in Bera would remain \\ strong and despite the harsh statement issued \\ by the top leaders of the two parties, priority \\ should be given to unite the Malay \\ Muslims and <MASK>.\end{tabular}} & Bumiputera & \multicolumn{1}{c|}{Christians} & \multicolumn{1}{c|}{Muslims} & \multicolumn{1}{c|}{Muslims} & \multicolumn{1}{c|}{,} & Muslims \\ \hline
\multicolumn{1}{|c|}{\begin{tabular}[c]{@{}c@{}}The Ministry of <MASK> Territories \\ (KWP) while ensuring the flood \\ management in Kuala Lumpur is \\ proceeding well.\end{tabular}} & Federal & \multicolumn{1}{c|}{New} & \multicolumn{1}{c|}{Protected} & \multicolumn{1}{c|}{Federal} & \multicolumn{1}{c|}{\#\#am} & Federal \\ \hline
\multicolumn{1}{|c|}{\begin{tabular}[c]{@{}c@{}}Hamzah said he had discussed the issue with \\ Inspector-General of Police Tan Sri \\ Acryl Sani <MASK> Sani.\end{tabular}} & Abdullah & \multicolumn{1}{c|}{-} & \multicolumn{1}{c|}{.} & \multicolumn{1}{c|}{Abdullah} & \multicolumn{1}{c|}{ser} & Abdullah \\ \hline
\multicolumn{1}{|c|}{\begin{tabular}[c]{@{}c@{}}Hamzah also said police had set up \\ a Tactical Command Centre in <MASK> \\ Langat district in Selangor to coordinate \\ flood relief operations of all units.\end{tabular}} & Hulu & \multicolumn{1}{c|}{the} & \multicolumn{1}{c|}{the} & \multicolumn{1}{c|}{Kuala} & \multicolumn{1}{c|}{,} & Hulu \\\hline
\end{tabular} 
}
\caption{Some sentences from the MEN-Dataset were used to predict masked tokens using various PLMs. Bold tokens indicate correctly predicted tokens when compared to the ground truth.}
\label{tab:random-mask}
\end{table*}

\citep{Salazar_2020, kauf2023better} suggests a "pseudo-log-likelihood" score calculated by masking tokens individually. The score is computed by summing the log-losses at the different masked positions. However, we are more interested in how accurately the models predict the masked token. This will help us to understand PLM models understanding and contextual awareness. We have collected 100 sentences from Malaysian English news article platform and we have done validation to ensure those sentence not part of our pre-training MEN-Corpus. In the first 70 sentences, one token from the local language, such as Bahasa Malaysia, has been randomly masked. In the remaining 30 sentences, one token of Standard English has been masked. We employ accuracy metrics to assess each model's effectiveness, providing a clear differentiation in their performance. The results demonstrate that additional pretraining on language-specific data significantly enhances the models' predictive capabilities, underscoring the importance of tailored training for improved language understanding. 

\begin{figure}[!ht]
    \centering
    \includegraphics[height=\textheight,width=0.5\textwidth,keepaspectratio]{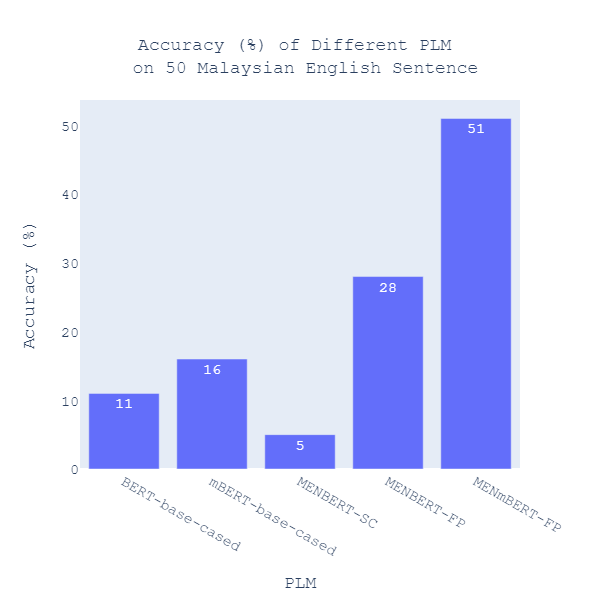}
    \caption{Accuracy of different pre-trained model predicting masked tokens in 50 Malaysian English Sentence.}
    \label{fig:accuracy_plm}
\end{figure}

For each pre-trained model, we calculated the accuracy of correctly predicted masked tokens. We used bert-based-cased and bert-base-multilingual-cased as baseline to investigate the improvement made by further pre-trained model. Based on the result in Figure \ref{fig:accuracy_plm}, we have identified that MENmBERT-FP has highest accuracy on predicting masked token from Malaysian English sentence. In Table \ref{tab:random-mask}, we present some sample of sentences showcasing how different PLMs perform in predicting masked tokens. Here are the findings obtained from the sample result shown in Table \ref{tab:random-mask}:
\begin{enumerate}
    \item MENmBERT-FP: Even though the model did not predict the exact ground-truth token, in some cases it identified semantically similar tokens. For instance, in the Malaysian context, \textit{Bumiputera} often refers to the \textit{Muslim} community. Here, MENmBERT-FP might predict a token related to ethnicity but not strictly synonymous with \textit{Muslim}. This highlights the model's ability to capture semantic nuances, even when encountering challenging cases. 
    \item MENBERT-FP: When we compared the performance of MENBERT-FP with bert-based-cased, we can understand that further-pretraining has improved the performance of language model. The success ratio of bert-based-cased is 0, and once further pre-trained, there is an improvement of +33\%.
    \item bert-based-cased: Since bert-based-cased has only been trained with English corpus, it was not able to unmask any tokens with compound blend correctly. 
    \item MENBERT-SC: Based on our observation, MENBERT-SC has produced bad results when unmasking the tokens. Once fine-tune, we will be able to understand better on the performance of the model. 
\end{enumerate}
In Section \ref{ssec: ft-result-and-analysis} we have discussed the performance of pre-trained model once fine-tune them for NER and RE.  

\begin{table*}[!ht]
\centering
\resizebox{\textwidth}{!}{%
\begin{tabular}{|c|cc|ccccc|}
\hline
Entity Label                                                              & \multicolumn{1}{c|}{\begin{tabular}[c]{@{}c@{}}Total Annotated \\ Entity in \\ MEN-Dataset\end{tabular}} & \begin{tabular}[c]{@{}c@{}}Total Annotated \\ Entity in \\ Validation Set\end{tabular} & \multicolumn{1}{c|}{bert-based-cased} & \multicolumn{1}{c|}{mbert-based-cased} & \multicolumn{1}{c|}{menbert-fp}     & \multicolumn{1}{c|}{menbert-sc}     & menmbert-fp    \\ \hline
PERSON                                                                    & \multicolumn{1}{c|}{1646}                                                                                & 108                                                                                    & \multicolumn{1}{c|}{0.74}             & \multicolumn{1}{c|}{0.84}              & \multicolumn{1}{c|}{0.79}           & \multicolumn{1}{c|}{0.17}           & \textbf{0.86}  \\ \hline
LOCATION                                                                  & \multicolumn{1}{c|}{1157}                                                                                & 150                                                                                    & \multicolumn{1}{c|}{0.86}             & \multicolumn{1}{c|}{0.88}              & \multicolumn{1}{c|}{0.87}           & \multicolumn{1}{c|}{0.48}           & \textbf{0.91}  \\ \hline
ORGANIZATION                                                              & \multicolumn{1}{c|}{1624}                                                                                & 262                                                                                    & \multicolumn{1}{c|}{0.81}             & \multicolumn{1}{c|}{\textbf{0.89}}     & \multicolumn{1}{c|}{0.82}           & \multicolumn{1}{c|}{0.29}           & \textbf{0.89}  \\ \hline
EVENT                                                                     & \multicolumn{1}{c|}{386}                                                                                 & 30                                                                                     & \multicolumn{1}{c|}{0.67}             & \multicolumn{1}{c|}{0.61}              & \multicolumn{1}{c|}{0.66}           & \multicolumn{1}{c|}{0.13}           & \textbf{0.77}  \\ \hline
PRODUCT                                                                   & \multicolumn{1}{c|}{72}                                                                                  & 6                                                                                      & \multicolumn{1}{c|}{0.24}             & \multicolumn{1}{c|}{\textbf{0.33}}     & \multicolumn{1}{c|}{0.13}           & \multicolumn{1}{c|}{0}              & 0.07           \\ \hline
FACILITY                                                                  & \multicolumn{1}{c|}{208}                                                                                 & 27                                                                                     & \multicolumn{1}{c|}{0.24}             & \multicolumn{1}{c|}{0.11}              & \multicolumn{1}{c|}{\textbf{0.47}}  & \multicolumn{1}{c|}{0}              & 0.25           \\ \hline
ROLE                                                                      & \multicolumn{1}{c|}{485}                                                                                 & 35                                                                                     & \multicolumn{1}{c|}{\textbf{0.39}}    & \multicolumn{1}{c|}{0.4}               & \multicolumn{1}{c|}{0.37}           & \multicolumn{1}{c|}{0.35}           & 0.6            \\ \hline
NORP                                                                      & \multicolumn{1}{c|}{114}                                                                                 & 5                                                                                      & \multicolumn{1}{c|}{0.88}             & \multicolumn{1}{c|}{0.6}               & \multicolumn{1}{c|}{\textbf{0.89}}  & \multicolumn{1}{c|}{0.21}           & 0.57           \\ \hline
TITLE                                                                     & \multicolumn{1}{c|}{300}                                                                                 & 4                                                                                      & \multicolumn{1}{c|}{\textbf{0.55}}    & \multicolumn{1}{c|}{0}                 & \multicolumn{1}{c|}{0.43}           & \multicolumn{1}{c|}{0.18}           & 0.5            \\ \hline
LAW                                                                       & \multicolumn{1}{c|}{62}                                                                                  & 5                                                                                      & \multicolumn{1}{c|}{0.15}             & \multicolumn{1}{c|}{0.12}              & \multicolumn{1}{c|}{\textbf{0.17}}  & \multicolumn{1}{c|}{0.1}            & 0.13           \\ \hline
LANGUAGE                                                                  & \multicolumn{1}{c|}{0}                                                                                   & 0                                                                                      & \multicolumn{1}{c|}{0}                & \multicolumn{1}{c|}{0}                 & \multicolumn{1}{c|}{0}              & \multicolumn{1}{c|}{0}              & 0              \\ \hline
WORK\_OF\_ART                                                             & \multicolumn{1}{c|}{7}                                                                                   & 2                                                                                      & \multicolumn{1}{c|}{0.1}              & \multicolumn{1}{c|}{0.15}              & \multicolumn{1}{c|}{0.12}           & \multicolumn{1}{c|}{0}              & \textbf{0.16}  \\ \hline
                                                                          & \multicolumn{1}{c|}{}                                                                                    &                                                                                        & \multicolumn{1}{c|}{}                 & \multicolumn{1}{c|}{}                  & \multicolumn{1}{c|}{}               & \multicolumn{1}{c|}{}               &                \\ \hline
Total Entities                                                            & \multicolumn{1}{c|}{6061}                                                                                & 634                                                                                    & \multicolumn{5}{l|}{}                                                                                                                                                       \\ \hline
\textbf{\begin{tabular}[c]{@{}c@{}}Overall\\ Micro F1-Score\end{tabular}} & \multicolumn{2}{c|}{\textbf{}}                                                                                                                                                                    & \multicolumn{1}{c|}{\textbf{0.763}}   & \multicolumn{1}{c|}{\textbf{0.819}}    & \multicolumn{1}{c|}{\textbf{0.778}} & \multicolumn{1}{c|}{\textbf{0.316}} & \textbf{0.831} \\ \hline
\end{tabular}%
}
\caption{Fine-Tuned model performance (based on F1-Score) calculated based on validation set for each entity labels. }
\label{tab:entity-label-plm}
\end{table*}

\begin{figure*}[!ht]
    \centering
    \includegraphics[height=\textheight,width=\textwidth,keepaspectratio]{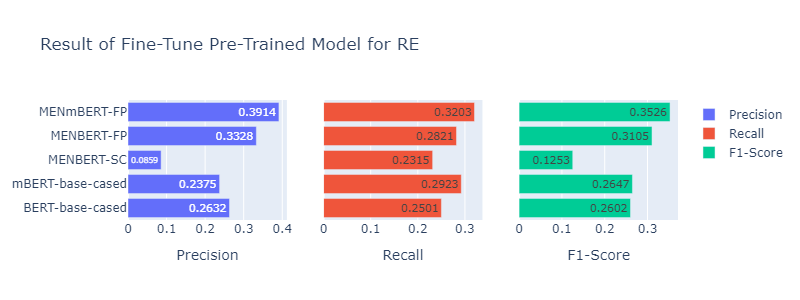}
    \caption{Precision, recall, and F1-score calculated for RE on the MEN-Dataset validation set.}
    \label{fig:Barchart-Evaluation-RE-FineTune}
\end{figure*}

\subsection{Fine-Tuning Pre-Trained Model}
\label{ssec: ft-result-and-analysis}

\subsubsection{Named Entity Recognition}
\label{sssec: ft-result-and-analysis-ner}
Figure \ref{fig:Barchart-Evaluation-NER-FineTune} shows the comparison of Precision, Recall, and F1-Score among five different pre-trained models. To evaluate the performance of further pre-trained models, we also fine-tuned pre-trained model (bert-base-cased, and bert-base-multilingual-cased) as the baseline. Meanwhile in Table \ref{tab:entity-label-plm}, we detailed the performance of model by entity labels.

Referring to the results, we observe that MENmBERT-FP achieves the highest F1-Score (0.831), while MENBERT-SC obtains the lowest F1-Score (0.316). Nevertheless, Figure \ref{fig:Barchart-Evaluation-NER-FineTune} demonstrates an improvement when further pre-training the BERT model, which validated our hypothesis (discussed in Section \ref{ssec: menbert-intro}). A few other observations from this experiment:
\begin{enumerate}
    \item MENmBERT-FP has a higher F1-Score (0.831) than mBERT-base-cased (0.819). We observe a +1.52\% improvement. Although the improvement is not significant, but when we analyse the F1-Score based on the entity label:
    \begin{enumerate}
        \item There is an significant improvement for 6 out of 11 entity labels that are evaluated. 
        \item On average, the difference in F1-Score between MENmBERT-FP and mBERT-base-cased is +10\%. 
        \item mBERT-base-cased is only able to achieve on-par in terms of F1-Score with MENmBERT-FP, specifically for entity label ORGANIZATION. 
    \end{enumerate}
    This in-depth observation proves the significant performance of MENmBERT-FP.  
    \item MENBERT-FP (F1-Score is 0.778) has a higher F1-Score than bert-based-cased (F1-Score is 0.763), with an improvement of +1.94\%. However, after further investigated the performance based on entity labels:
    \begin{enumerate}
        \item MENBERT-FP has only improved the performance of only 7 out of 11 entity labels. 
        \item On average, the improvement is only around +2\%.
        \item When compared with MENmBERT-FP, MENBERT-FP is able to outperform in terms of F1-Score for 4 out of 11 entity labels. These include entity labels PRODUCT, FACILITY, NORP and LAW. 
    \end{enumerate}
    \item MENBERT-SC has not shown any improvement in the performance of NER. Our evaluation in Section \ref{ssec: fp-result-and-analysis} also shows it was not able to unmask the tokens correctly.  
\end{enumerate}
The experimental results and findings conclude that our fine-tuned MENmBERT-FP has achieved highest F1-Score compared to other pre-trained models. MENmBERT-FP could be used to fine-tune for more NLP downstream tasks, involving Malaysian English, for improved performance.

\subsubsection{Relation Extraction}
\label{sssec: ft-result-and-analysis-re}
Figure \ref{fig:Barchart-Evaluation-RE-FineTune} shows F1-Scores compared across five PLMs. Like with NER, we used fine-tuned bert-base-cased and bert-base-multilingual-cased as baselines. 

MENmBERT-FP achieved the highest F1-score (0.353), indicating a slight improvement over our baseline PLMs. This suggests that further pre-training on MEN-Dataset has been beneficial for RE on Malaysian English context. Here are some of ur observations from the experiment:
\begin{enumerate}
    \item MENmBERT-FP has made +33.21\% improvement in F1-Score compared to baseline approach mBERT-base-cased. This has been proven significant. Our cross-analysis of NER and RE predictions reveals that the TP relation instances have entity pairs correctly classified by the fine-tuned NER model (from previous analysis). This suggests a significant improvement in RE performance due to the model's enhanced entity prediction capabilities. 
    \item MENBERT-FP (F1-Score is 0.3105) has a higher F1-Score than bert-based-cased (F1-Score is 0.2602), with an improvement of +19.33\%. The analysis of the fine-tuned model's predictions did not reveal any surprising or unexpected patterns. This aligns with the observations from the previous point. Meanwhile, for MENBERT-SC was performed badly when fine-tuned for RE task.
\end{enumerate}
Apart from that, it's important to note that our fine-tuned RE models achieved lower performance compared to reported results on other document-level relation extraction datasets like DocRED. For instance, prior work using a fine-tuned BERT model on DocRED (38,269 relation instances) achieved an F1-score of 54.16 (Dev) and 53.20 (Test) \citep{wang2019finetune}. Compared with our finding, the overall F1-Score could be not significant due to the nature of MEN-Dataset with a small set of annotation instance. 

\section{Conclusion}
\label{sec:conclusion-future-work}
This work introduced MENmBERT, a contextualized language model pre-trained on a Malaysian English corpus. Our experiments demonstrated that fine-tuning MENmBERT on language-specific data significantly improves performance on NER tasks with average of +1.74\%. For RE, we have achieved average improvement of +26.27\% compare our MENmBERT and MENBERT with baseline PLM's. However, the fine-tuned RE models achieved lower performance compared to reported results on other document-level relation extraction datasets. This suggests that while MENmBERT's entity prediction capabilities benefit RE tasks, further exploration is needed to optimize RE performance in the context of our dataset. This has gaps has suggested us for future work, explore several avenues to improve RE performance. One of the approach involve investigating data augmentation techniques to expand our dataset and improve model training. Beyond that, we will extend our experiment do several other downstream NLP task.

\section{Acknowledgements}
\label{sec:ack}
Part of this project was funded by the Malaysian Fundamental Research Grant Scheme (FRGS) FRGS/1/2022/ICT02/MUSM/02/2.
\bibliography{custom}

\onecolumn
\appendix

\end{document}